\documentclass[conference]{IEEEtran}
\IEEEoverridecommandlockouts

\usepackage{graphicx}
\usepackage{amsmath, amssymb}
\usepackage{cite}
\usepackage{lipsum}
\usepackage{multirow}
\usepackage{caption}
\usepackage{float}
\usepackage{booktabs}
\usepackage{balance}
\usepackage{hyperref}

\begin{document}

\title{Large-scale Photorealistic Outdoor 3D Scene Reconstruction from UAV Imagery Using Gaussian Splatting Techniques
\thanks{This work has received funding from the European Union’s Horizon Europe research and innovation programme under Grant Agreement No. 101168042 project TRIFFID (auTonomous Robotic aId For increasing First responders Efficiency). The views and opinions expressed in this paper are those of the authors only and do not necessarily reflect those of the European Union or the European Commission.}
}

\author{
\IEEEauthorblockN{
Christos Maikos\IEEEauthorrefmark{1},
Georgios Angelidis\IEEEauthorrefmark{1},
Georgios Th. Papadopoulos\IEEEauthorrefmark{1}\IEEEauthorrefmark{2}
}

\IEEEauthorblockA{\IEEEauthorrefmark{1}
Department of Informatics and Telematics, Harokopio University of Athens, Athens, Greece
}

\IEEEauthorblockA{\IEEEauthorrefmark{2}
Archimedes, Athena Research Center, Athens, Greece \\
Email: \{chmaikos, gangelidis, g.th.papadopoulos\}@hua.gr}
}

\maketitle

\begin{abstract}
In this study, we present an end-to-end pipeline capable of converting drone-captured video streams into high-fidelity 3D reconstructions with minimal latency. Unmanned aerial vehicles (UAVs) are extensively used in aerial real-time perception applications. Moreover, recent advances in 3D Gaussian Splatting (3DGS) have demonstrated significant potential for real-time neural rendering. However, their integration into end-to-end UAV-based reconstruction and visualization systems remains underexplored. Our goal is to propose an efficient architecture that combines live video acquisition via RTMP streaming, synchronized sensor fusion, camera pose estimation, and 3DGS optimization, achieving continuous model updates and low-latency deployment within interactive visualization environments that supports immersive augmented and virtual reality (AR/VR) applications. Experimental results demonstrate that the proposed method achieves competitive visual fidelity, while delivering significantly higher rendering performance and substantially reduced end-to-end latency, compared to NeRF-based approaches. Reconstruction quality remains within 4-7\% of high-fidelity offline references, confirming the suitability of the proposed system for real-time, scalable augmented perception from aerial platforms.
\end{abstract}


\section{Introduction}

UAV-based applications are constantly increasing, as modern drones equipped with cameras, sophisticated sensors and artificial intelligence modules are widely adopted to provide cheap and effective solutions for a broad range of tasks, which vary from agricultural tasks \cite{agrawal2024}  \cite{Manoj2025} to industry \cite{ejaz2024, alimisis2025advances, bright2025} and pre- and post-disaster assessments \cite{BOROUJENI2024102369, cani2025}. The establishment of systems capable of processing live video captured from UAVs and converting them into realistic 3D visualisations in real-time constitutes a research field that combines computer vision, computer graphics, and interactive visualisation techniques. 

Drone usage for data collection supports distinct adaptability in terms of coverage and perspective, allowing the acquisition of rich, multi-layered and high-resolution information that encompasses both motion dynamics and environmental characteristics \cite{shukla24}. 
The acquired information can be supplemented by utilising photogrammetry tools designed for 3D structure recording, where the fusion of spectral channels enables enhanced separation between foreground and background.

\begin{figure}[t]
    \centering
    \includegraphics[width=0.95\columnwidth]{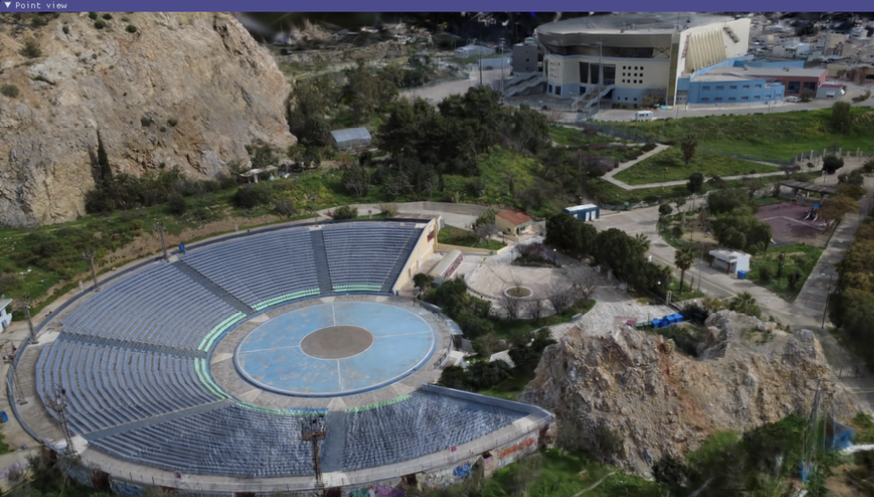}
    \caption[width=0.8\columnwidth]{Complex stadium site reconstructed using the proposed approach.}
    \label{fig:katrakeio}
\end{figure}

During recent years, Gaussian Splatting has offered considerable improvements regarding the quality of real-time scene rendering \cite{Pang_2024_CVPR}, while recent approaches consider dynamic parameterisation for time-varying scenes or objects \cite{Qian_2024_CVPR}. One major benefit of this representation is the ability to bind and work with deformable meshes using UV texture space mapping \cite{Lupiani2025} to improve performance even in complex motion conditions.
Moreover, the integration of 3DGS \cite{Kerbl2023} into a representation engine environment offers the ability to directly interact with the 3D entities, even in mixed or AR-based applications. Nevertheless, this process requires compatibility with various rendering and representation engines, such as WebGL \cite{CHICKERUR2024919} and Unity \cite{computers15010068}, as well as sophisticated rendering optimisations, even in resource-limited devices, such as AR glasses.

Visual quality perception depends not only on the data acquisition stage but also on continuous processing throughout the acquisition and rendering phases, due to the quality loss caused by compression. Moreover, data transmission should be optimized considering the network constraints. Systems based on visual spectrum and depth information stream transmission can leverage established high-fidelity streaming methods even on edge devices, provided compression formats remain adaptive \cite{gunkel2023}.

In poor signal or low bandwidth conditions, transmitting the entirety of a 3D model in real-time may prove difficult \cite{alexovivc20233d}, highlighting the importance of the selective transmission of critical information or sampling rate adaptation. Recent advances in projecting Gaussians onto a 2D image plane have created new perspectives in efficiency improvement. Through transformations in SE(3) space, rapid mapping of points to camera space, and subsequently to the image one, is achievable \cite{barad2024}, a process critical when input originates from multiple observation points.

\begin{figure*}[t]
    \centering
    \includegraphics[width=0.8\textwidth]{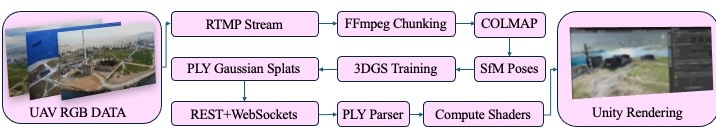}
    \caption{Developed pipeline for converting UAV video streams into 3DGS representations integrated into a Unity environment.}
    \label{fig:system_pipeline}
\end{figure*}

In tandem, efforts to integrate semantic features \cite{cani2026illicit} into each Gaussian strengthen the analysis of complex scenes \cite{Zhou_2024_CVPR}, a property that drastically facilitates detection and interaction with distinct objects within them. The objective of using such a system for augmented environmental perception requires rigid temporal consistency among frames for the visualisation to appear realistic to the user or to the trajectory of the drone. Previous implementations have demonstrated that efficient Gaussian renderers are capable of delivering high frame rates without sacrificing visual quality \cite{Barthel_2024_CVPR}, which is a critical aspect when dealing with VR/AR systems, as latency above a few milliseconds is noticeable to the user.

Considering the above, we propose a comprehensive pipeline that leverages 3DGS to convert live UAV video streams into high-fidelity 3D scenes with minimal latency. Our novelties include the following:

\begin{itemize}
    \item Real-Time 3DGS Reconstruction: We present a system capable of processing live UAV footage into geometrically consistent 3D Gaussian representations.
    \item Seamless AR/VR Integration: The proposed architecture integrates directly with the visualisation engine, enabling interactive, immersive visualization and AR applications.
    \item Adaptive Streaming Architecture: By utilizing RTMP for data collection and WebSockets for real-time updates, the system dynamically adapts to network conditions and efficiently supports resource-constrained devices.
\end{itemize}

The remaining sections of this paper are structured as follows: Section \hyperref[sec:related-work]{2} provides information on existing work related to 3DGS reconstruction. Section \hyperref[sec:methodology]{3} introduces our system’s background theory and methodology. Section \hyperref[sec:experiments]{4} presents our evaluation approach and setup. Section \hyperref[sec:results-and-discussion]{5} discusses the obtained results and finally, our conclusion and possible expansions of our work are provided in Section \hyperref[sec:conclusion]{6}.

\section{Related Work} \label{sec:related-work}

Dynamic scene representation is a constantly evolving field, especially in the latest years where the neural scene representations have been widely adopted in the bibliography. In what follows, an analysis of relevant studies is presented regarding neural 3DGS-based scene reconstruction.

In this context, Tang et al. \cite{Tang_2025_CVPR_droneSplat} proposed a 3DGS framework for aerial imagery that utilizes adaptive masking and voxel-guided optimization to handle dynamic distractors and sparse viewpoints. Similarly, 
Shang et al. \cite{Shang_3DGS_mdpi} introduced an enhanced 3DGS framework that leverages metric depth priors, gradient-driven adaptive densification, and density-based artifact pruning to improve reconstruction accuracy and efficiency in complex real-world scenes. 
Ham et al. \cite{Ham_2024_Dragon_3DGS} proposed a framework enabling robust 3D reconstruction from  drone and ground imagery by iteratively extrapolating intermediate viewpoints with perceptual regularization to bridge the visual feature gap preventing standard registration, while 
Wu et al. \cite{wu_2025_flyGS} developed a method tailored for large-scale aerial reconstruction that utilizes a hybrid Gaussian representation with a dual-stage optimization strategy to balance efficient rendering speed with high-fidelity geometric detailing. In the same line, 
Luo et al. \cite{Luo_2026_mdpi_3DGS} introduced a depth-cue aware framework that refines geometric initialization using a minimum spanning tree algorithm and employs progressive depth-guided training to manage the significant depth variations inherent in drone imagery. 
Qian et al. \cite{Qian_2025_compressed_3DGS} introduced a framework for large-scale UAV that prunes redundant Gaussians based on a contribution score and employing ray-tracing volume rendering to preserve fine geometric details.
Mei et al. \cite{mei_2025_3DGS_lowatt_UAV} developed a scalable divide-and-conquer framework for urban reconstruction that integrates distributed Structure-from-Motion (SfM) with dense depth priors to drive parallelized 2DGS, effectively mitigating memory constraints while preserving boundary consistency through a dedicated refinement strategy.

The proposed system effectively bridges the gap between aerial data acquisition and immersive visualisation by integrating 3DGS into a real-time visualization environment. Unlike optimization-centric frameworks, such as \cite{Tang_2025_CVPR_droneSplat, Shang_3DGS_mdpi}, which primarily focus on handling dynamic distractors or leveraging depth priors to enhance reconstruction accuracy, our approach prioritizes the efficiency of the end-to-end pipeline. Furthermore, while recent hybrid approaches \cite{Ham_2024_Dragon_3DGS, mei_2025_3DGS_lowatt_UAV} successfully merge ground and aerial level imagery for geometric completeness, they often lack the direct compatibility with representation engines required for immediate AR applications, a capability that is considered crucial for our method.

\section{Methodology}
\label{sec:methodology}

Figure \ref{fig:system_pipeline} depicts the pipeline of the proposed system. The system converts UAV-acquired video streams and sensor data, into geometrically and temporally consistent, semantically enriched inputs suitable for SfM/Multi-View Stereo (MVS) and 3DGS, while maintaining low end-to-end latency.

\subsection{Principles of 3D Gaussian Representations} \label{sec:principles-of-3dgs}

3DGS represents a scene as a set of anisotropic Gaussian distributions in a 3D space. Each Gaussian is described by a mean position $\mu$ and covariance $\Sigma$. During rendering, these 3D Gaussians are projected through the camera model into the image plane, as 2D Gaussian splats whose size, shape, and orientation depend on the viewing configuration. Appearance is modelled using spherical harmonics coefficients attached to each primitive. An opacity parameter controls the contribution of each Gaussian to the rendered image, while training is performed by minimizing a photometric loss between rendered images and ground-truth frames. 
New Gaussians can be injected in regions with high reconstruction error or that become visible later in a sequence while existing Gaussians can be modified or removed in an evolving scene. For dynamic scenes, they can be associated with non-static meshes and animated, considering local lighting effects. Moreover, they can carry semantic information allowing the renderer to output not just RGB images but also semantic or instance maps, enabling rich AR interactions.

\subsection{Data collection and RTMP server integration} \label{sec:data-collect}
The proposed architecture is built around a data collection subsystem that receives live streams from UAVs equipped with RGB-D or multispectral cameras. Each robot uses hardware-accelerated H.264/H.265 encoding to compress the video, to reduce flight controller's load, and to transmit it over a reliable RTMP channel. Control, telemetry and video data are transmitted over separate channels to avoid congestion, especially from heavy I-frames that could delay flight control packets. In multi-UAV scenarios, each agent publishes its own RTMP stream to a common media server, capable of sustaining multiple persistent TCP connections. Buffer sizes, queuing, and retransmission parameters are tuned to achieve end-to-end latencies of a few hundred milliseconds.

On the server side, incoming RTMP streams are decoded into raw frames, which can be passed directly to processing pipelines via shared memory or IPC mechanisms. At this stage, the system may also perform adaptive bitrate or resolution adjustments based on backend load and network conditions: when the reconstruction pipeline becomes saturated or bandwidth drops, the server can dynamically reduce the outgoing resolution or bitrate to maintain responsiveness. Frames are fed into separate processing paths for reconstruction, direct AR overlay, and preview, preventing computationally heavy tasks such as dense reconstruction from blocking immediate visual feedback. Overall, the RTMP server is a vital component of a broader data integrity and latency management strategy that strongly influences the final visual quality and temporal consistency of the AR experience. Once a reconstruction session completes, its output is delivered to client-side visualization and interaction applications through a real-time bidirectional communication interface. This push-based mechanism ensures that rendering engine applications automatically receive the latest map representations. Clients choose how to handle updates in near real-time: replace their current model, merge the new splats with existing ones, or selectively loading regions of interest.

\subsection{Frame extraction and synchronization} \label{sec:frame-extr}
The system processes compressed multimedia streams and decodes them into raw image buffers, assigning timestamps based on stream metadata. To ensure consistency across multimodal inputs, each stream is normalized to a common time base, using network synchronization protocols. A dedicated synchronization module maintains separate queues for each modality, aligning video frames with the nearest corresponding sensor samples within a configurable temporal window. Missing sensor data is reconstructed via interpolation for linear metrics or integration for inertial measurements, ensuring a continuous and complete data stream.

To manage processing load, frame rate reduction is applied either uniformly or adaptively, selecting frames based on motion magnitude or segmentation dynamics, while preserving full frame sequences for high-speed content to capture fine temporal details. Addressing the inherent latency disparity between buffered video streams and real-time telemetry, the architecture employs a multi-threaded buffering system that aligns streams based on provided timestamps. This approach decouples decoding from synchronization, ensuring the strict temporal coherence necessary for geometrically accurate alignment in the final reconstruction pipeline.

\subsection{Camera pose estimation} \label{sec:cam-pose-est}
Camera pose estimation converts temporally aligned sensor data into 6-DoF poses, defining the position and orientation of each camera frame in a global reference frame. RGB-D data allows visual odometry and SLAM, while monocular RGB uses SfM pipelines with feature matching, geometric verification, and bundle adjustment; additional sensors, IMU/GPS data, and semantic segmentation help stabilize poses in challenging UAV scenarios. The output is a sequence of SE(3) transformations 
\[
T_i = [\,R_i \mid t_i\,],
\] 
where $R_i$ is the $3 \times 3$ rotation matrix and $t_i$ is the 3D translation vector of camera $i$ in the global coordinate system. These poses are used by the renderer to project 3D Gaussians into each view without spatial or temporal discontinuities.

\subsection{Training and deployment of the 3DGS model} \label{sec:train-and-deploy}
The training procedure initializes 3D Gaussians at point locations derived from the SfM/MVS cloud, defining each primitive as described in Section \ref{sec:principles-of-3dgs}. Frames are processed in mini-batches to maximize GPU utilization. Parameters are optimized via a differentiable tile-based rasterizer that projects Gaussians into camera views to compute photometric loss against ground-truth frames. This process incorporates an adaptive density control mechanism, where a densification phase involves sampling new points in under-reconstructed regions, coupled with a pruning stage that removes low-contributing primitives, ensuring model compactness, geometric integrity, and mathematical accuracy. For deployment, the system employs mixed-precision training and spatial regularization to prevent artifacts. The final model is stored in a compact binary format containing contiguous parameter arrays and spatial tiling data. In live deployments, the model is continuously updated as new frames arrive. Instead of retraining from scratch, the system performs online optimization focused on regions affected by the new data, achieving a continuously updated and spatially consistent 3DGS representation.

\section{Experimental Evaluation}\label{sec:experiments}

\subsection{Datasets}\label{sec:datasets}

We evaluated our method on three widely used multi-view reconstruction and novel view synthesis benchmarks: Mip-NeRF 360 \cite{Barron_2022_mipNerf360}, Tanks and temples \cite{Knapitsch2017}, and Deep blending \cite{DeepBlending2018}. Combined, the selected benchmarks cover a diverse set of challenges, including scene scale, geometry complexity, and photorealistic view synthesis.

\subsection{Experimental Setup}
The experimental setup is designed with an emphasis on the pipeline behaviour on real-world information. The main goal is to characterize the end-to-end performance of the backend. UAVs capture RGB or RGB-D video alongside IMU and GPS telemetry, which is streamed to a ground station via RTMP \cite{RTMP-reference} and decoded in near real-time. To capture realistic networking and buffering behavior, prerecorded sequences can be re-streamed using the NGINX RTMP module \cite{nginxServer-reference}. Video and sensor streams share a unified time base, namely IEEE 1588 PTP \cite{ieee1588PTP-reference}. Reconstruction updates are distributed to clients over WebSocket (RFC 6455) \cite{webSocket-reference}, enabling measurement of network delivery latency and client update behavior. The backend executes on a workstation equipped with an AMD Ryzen 9 9900X3D, an NVIDIA RTX 4070 12 GB, and 64 GB RAM. The software pipeline integrates FFmpeg decoding, python-based preprocessing and semantic segmentation, COLMAP-style SfM/MVS \cite{colmap-reference}, and an original INRIA 3DGS implementation \cite{Kerbl2023}. We evaluated two variants of our method, different in the number of Gaussians used for the reconstruction: Ours30K with 30,000 iterations and Ours7K with 7,000 iterations. Unity \cite{computers15010068} was selected as the visualisation engine and client, while the utilized drone was a DJI Mini 4K.

\subsection{Evaluation metrics}
The evaluation process addresses visual fidelity, geometric accuracy, and operational performance. Visual fidelity on held-out views is quantified via PSNR, SSIM \cite{Wang04}, and LPIPS \cite{Zhang_2018_CVPR}, computed cumulatively to monitor drift. We adopt accuracy and completeness reporting as in \cite{schops-2017-cvpr}. When using Tanks and Temples, we report a distance-threshold F-score adopted from \cite{Knapitsch2017}. Pose accuracy is evaluated via Absolute Trajectory Error (ATE) and Relative Pose Error (RPE) following  \cite{strum_12}. Finally, semantic quality is measured using mIoU and, for instance-aware tasks, Panoptic Quality (PQ) \cite{Kirillov_2019_CVPR}. The operational metrics also include: end-to-end latency, throughput and update stability.

\subsection{Results}

Table \ref{table:bench-comp} reports the quantitative performance of Instant-NGP \cite{muller_2022_instantNGP}, Mip-NeRF360 \cite{Barron_2022_mipNerf360}, and the proposed 3DGS method variants across the selected benchmarks. On the Mip-NeRF360 dataset, our proposed Ours30K algorithm achieves an SSIM of 0.815, a PSNR of 27.21, and an LPIPS of 0.214, with a total training time of 41 minutes and a rendering speed of 134 FPS. The same method, when evaluated on Tanks and temples, records an SSIM of 0.841, PSNR of 23.14, and LPIPS of 0.183, while maintaining a rendering speed of 154 FPS. Similarly, for the Deep blending dataset, it reaches an SSIM of 0.903, PSNR of 29.41, and LPIPS of 0.243, with a training time of 36 minutes and a rendering speed of 137 FPS.
The Ours7K configuration exhibits reduced training time across all datasets, while maintaining comparable image quality metrics and higher rendering speeds. Instant-NGP demonstrates lower memory usage and shorter training times, whereas Mip-NeRF360 presents higher training costs with reduced rendering speed across all evaluated datasets.

\begin{table*}[t] 
\centering
\caption{Experimental evaluation and comparison of reconstruction quality and resource efficiency.}
\label{table:bench-comp}
\label{tab:quantitative_results}
\begin{tabular}{l l c c c c c c}
\hline
\textbf{Dataset} & \textbf{Method} & \textbf{SSIM}  & \textbf{PSNR}  & \textbf{LPIPS}  & \textbf{Train}  & \textbf{FPS}  & \textbf{Memory}  \\
\hline
\multirow{4}{*}{Mip-NeRF360 \cite{Barron_2022_mipNerf360}}
& Instant-NGP (Base) & 0.671 & 25.30 & 0.371 & 5m 37s & 11.7 & 13MB \\
& Mip-NeRF360 & 0.792 & 27.69 & 0.237 & 48h & 0.06 & 8.6MB \\
& Ours7K & \textbf{0.770} & \textbf{25.60} & \textbf{0.279} & \textbf{6m 25s} & \textbf{160} & 523MB \\
& Ours30K & 0.815 & 27.21 & 0.214 & 41m 33s & 134 & 734MB \\
\hline
\multirow{4}{*}{Tanks and temples \cite{Knapitsch2017}}
& Instant-NGP (Base) & 0.723 & 21.72 & 0.330 & 5m 26s & 17.1 & 13MB \\
& Mip-NeRF360 & 0.759 & 22.22 & 0.257 & 48h & 0.14 & 8.6MB \\
& Ours7K & \textbf{0.767} & \textbf{21.20} & \textbf{0.280} & \textbf{6m 55s} & \textbf{197} & 270MB \\
& Ours30K & 0.841 & 23.14 & 0.183 & 26m 54s & 154 & 411MB \\
\hline
\multirow{4}{*}{Deep blending \cite{DeepBlending2018}}
& Instant-NGP (Base) & 0.797 & 23.62 & 0.423 & 6m 31s & 3.26 & 13MB \\
& Mip-NeRF360 & 0.901 & 29.40 & 0.245 & 48h & 0.09 & 8.6MB \\
& Ours7K & \textbf{0.875} & \textbf{27.78} & \textbf{0.317} & \textbf{4m 35s} & \textbf{172} & 386MB \\
& Ours30K & 0.903 & 29.41 & 0.243 & 36m 2s & 137 & 676MB \\
\hline
\end{tabular}
\end{table*}

\section{Discussion} \label{sec:results-and-discussion}
The proposed system leverages 3DGS, achieving improved rendering speeds compared to NeRF-based algorithms that rely on computationally expensive ray marching techniques. While view-synthesis quality is comparable to improved NeRFs, 3DGS is notably faster in training and rendering. Unlike point clouds, anisotropic Gaussian kernels minimize aliasing and geometry gaps, and the addition of spherical harmonics captures view-dependent lighting often lacking in standard MVS pipelines.

For dynamic scenes, attaching Gaussians to deformable objects via UV mapping exhibits superior performance compared to traditional linear blend skins by accurately reproducing precise details, without requiring extensive model retraining. Additionally, when compared to voxel-based representations, 3DGS scales linearly, utilizing tile-based rasterization to handle large UAV datasets without bottlenecks. Furthermore, the nature of splats allows for local splat rearrangements and frame-by-frame updates without the global side effects characterising NeRFs.

Operational efficiency is maintained through progressive loading and compact parameter codebooks, facilitating this way the deployment on resource-constrained immersive devices. The Unity integration achieves continuous asset streaming and supports multi-user collaboration. Low-latency can be accomplished by leveraging RTMP/RTMPS combined with hardware acceleration and telemetry channels that manage network constraints effectively.

The developed system remains sensitive to input data quality. Potential camera pose errors can propagate cumulatively, causing visual artifacts. Moreover, while dynamic updates and depth-based occlusion enhance realism, they notably stress the GPU. Additionally, pre-processing relies heavily on deep learning accuracy, where segmentation errors can burden the SfM pipeline. Despite these challenges, quantitative evaluations indicate the method maintains reconstruction quality within 4-7\% of a high-fidelity offline reference, while substantially reducing end-to-end latency. This trade-off enables real-time, scalable augmented perception and outperforms foreground-mixing approaches such as those in \cite{bulinger2018}.

\section{Conclusion} \label{sec:conclusion}
This work presents a comprehensive approach to the real-time reconstruction of 3D scenes leveraging UAV-acquired data, based on 3DGS. This method brings together high geometric precision with computational efficiency and has the potential of being utilised in real-time AR applications. The explicit use of Gaussian representations yields numerous advantages over alternative approaches, specifically regarding reduced rendering times and the capability for local model updates without disrupting the global scene context. The architecture of the system develops an integral technological infrastructure that encompasses UAV data capture and processing, up to AR display that ensures a continuous information flow with minimal latency.

Scalability across multiple streams and support for computationally-constrained devices render this system extremely flexible and applicable in various environments and operational scenarios. Despite the discussed limitations and challenges, such as camera pose estimation accuracy and dynamic content management, the proposed method outperforms other techniques in terms of both efficiency and visual quality. Its deployment in applications such as archaeological documentation, surveillance, or remote collaboration opens a whole dimension of possibilities addressing practical problems. Future work will focus on optimizing input data accuracy, integrating state-of-the-art machine learning algorithms, and extending capabilities to even demanding environments. Ultimately, the value of this approach lies in its ability to merge advanced technological performance with practical utility, providing a robust tool for augmented perception and real-world interaction. Moreover, the system is designed so as to be connected with real-time AI-enabled event detectors \cite{linardakis2025survey, foteinos2025visual, linardakis2024distributed, konstantakos2025self}, so as to facilitate scenarios involving Human-Robot Interaction \cite{papadopoulos2021towards, papadopoulos2022user, moutousi2025tornado} and broader security response incidents \cite{mademlis2024invisible}, while also supporting the necessary eXplainable Artificial Intelligence (XAI) pipelines \cite{rodis2024multimodal, evangelatos2025exploring}.

\balance

\bibliographystyle{IEEEtran}
\bibliography{references}

\end{document}